\documentclass{article}

\usepackage{caption}
\usepackage{subcaption}

\usepackage{microtype}
\usepackage{graphicx}
\usepackage{booktabs} %

\usepackage{hyperref}

\usepackage[accepted]{icml2025}

\usepackage{amsmath}
\usepackage{amssymb}
\usepackage{mathtools}
\usepackage{amsthm}

\usepackage[capitalize,noabbrev]{cleveref}

\theoremstyle{plain}

\theoremstyle{definition}

\theoremstyle{remark}

\usepackage[textsize=tiny]{todonotes}

\usepackage{setspace}

\icmltitlerunning{ Token-level uncertainty and hidden state dynamics in language models }

\begin{document}

\twocolumn[
\icmltitle{Are language models aware of the road not taken? \\ Token-level uncertainty and hidden state dynamics}

\begin{icmlauthorlist}
\icmlauthor{Amir Zur}{stanford,prair}
\icmlauthor{Atticus Geiger}{goodfire,prair}
\icmlauthor{Ekdeep Singh Lubana}{ntt,hcbs}
\icmlauthor{Eric Bigelow}{ntt,hpsy}
\end{icmlauthorlist}

\icmlaffiliation{prair}{Pr(Ai)$^2$R Group}
\icmlaffiliation{goodfire}{Goodfire}
\icmlaffiliation{stanford}{Department of Linguistics, Stanford University}
\icmlaffiliation{hpsy}{Department of Psychology, Harvard University}
\icmlaffiliation{hcbs}{Center for Brain Science, Harvard University}
\icmlaffiliation{ntt}{Physics of Intelligence Group, NTT Research}

\icmlcorrespondingauthor{Amir Zur}{amirzur@stanford.edu}
\icmlcorrespondingauthor{Eric Bigelow}{ebigelow@g.harvard.edu}

\icmlkeywords{Machine Learning, ICML}

\vskip 0.3in
]

\printAffiliationsAndNotice{}  %

\begin{abstract}
When a language model generates text, the selection of individual tokens might lead it down very different reasoning paths, making uncertainty difficult to quantify.
In this work, we consider whether reasoning language models represent the alternate paths that they could take during generation.
To test this hypothesis, we use hidden activations to control and predict a language model's uncertainty during chain-of-thought reasoning. 
In our experiments, we find a clear correlation between how uncertain a model is at different tokens, and how easily the model can be steered by controlling its activations. This suggests that activation interventions are most effective when there are alternate paths available to the model---in other words, when it has not yet committed to a particular final answer.
We also find that hidden activations can predict a model's future outcome distribution, demonstrating that models implicitly represent the space of possible paths. 
\end{abstract}

\section{Introduction}
\label{sec:intro}

In recent years, Large Language Models (LLMs) have shown impressive capabilities which emerge during next-word prediction \citep{brown2020language}. 
Despite this high performance on many intelligence benchmarks, LLMs often fail in unexpected and sometimes dramatic ways, for example confidently hallucinating wrong answers or outputting harmful text.
Under the surface, when an LLM samples a sequence of text to output, at each token there is a possibility that the LLM might seemingly ``change its mind'' and say something very different.
The goal of this work is to study the latent neural representations underlying these token-level uncertainty dynamics, inspired by work evaluating the abilities of LLMs and interpreting their inner workings \citep{templeton2024scaling, panickssery2023steering, ferrando2024primer}.

\begin{figure}[t]
    \centering
    \includegraphics[width=\linewidth]{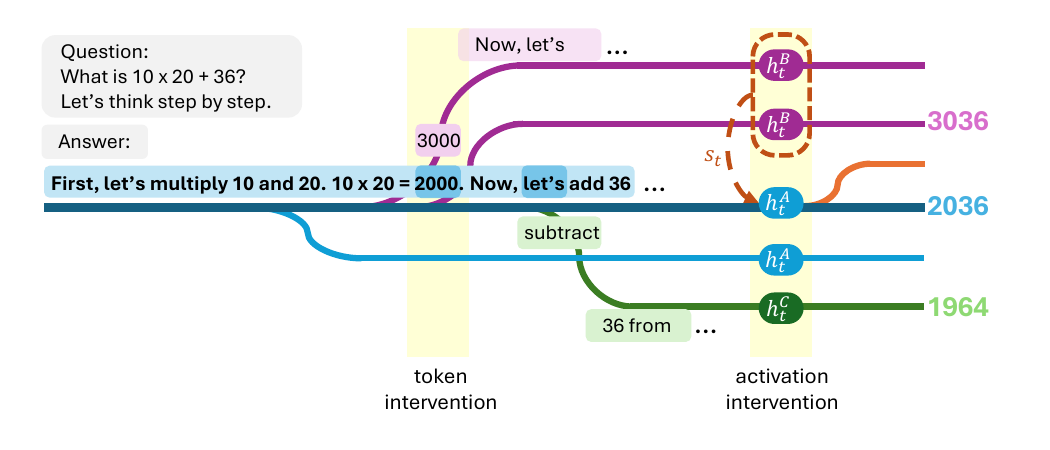}
    \caption{Our experimental set-up. By intervening on the generated tokens, we create branching paths to estimate the model's outcome distribution. By intervening on the model's activations, we steer the base generation towards a desired outcome.}
    \label{fig:main}
    \vspace{-10pt}
\end{figure}

Uncertainty estimation is a foundational problem in evaluating and interpreting LLMs \citep{geng2024survey, xiong2024can}. For example, if an LLM ``hallucinates'' a wrong answer to a question, but the probability of this answer is only 1\%, this LLM should be evaluated differently than another LLM that outputs the same wrong answer with 100\% confidence.
However, the challenge of estimating uncertainty in LLMs is considerably more difficult in settings with long-form text responses, such as with reasoning models where a chain of reasoning text precedes the final answer \citep{kojima2022large, jaech2024openai, guo2025deepseek}. For each token that an LLM generates, sampling a different token might lead the LLM to generate a different answer.
In particular, \citet{bigelow2025forking} develop an approach, called Forking Path Analysis, to demonstrate the important role that individual tokens play in determining model certainty.

There is much recent work in AI interpretability which seeks to understand the latent representations in LLMs, and to develop methods for steering or controlling LLMs by intervening on hidden representations.
One common approach to interpretability is to study a particular phenomenon in isolation, such as induction heads \citep{olsson2022context}, with small autoregressive language models trained on toy problems such as sequences of characters or functional input--output pairs \citep{akyurek2022learning, xie2021explanation}.
An alternative approach to studying LLMs is to instead study in-context learning dynamics and representation learning in pre-trained LLMs, at the level of individual tokens \citep{park2025iclr, bigelow2023context}.
While these approaches emphasize the role of in-context learning and autoregressive text generation, these aspects of text generation are typically not considered in research on steering. Steering methods strive to construct interventions on hidden representations, such as a vector representation of a concept which can be added to the residual stream of a transformer LLM to steer its behavior \citep{marks2023geometry,li2023inference,turner2023steering,rimsky2024steering}. For example, \citet{arditi2024refusal} construct a vector that makes an LLM more or less likely to refuse answering a user's question. \citet{fei2024nudging} develop a method for steering an LLM at the token level, using a smaller LLM as a guide to intervening by replacing specific tokens with alternate strings that lead the LLM down a different reasoning path.

\setlength{\fboxsep}{1pt}

\begin{figure}[t]
    \centering
    \includegraphics[width=0.95\linewidth]{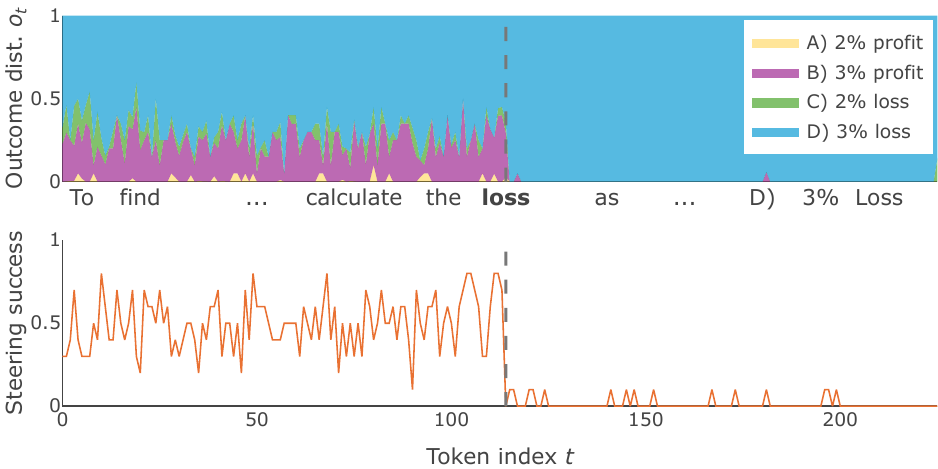}
    \vskip 5pt
    \begin{minipage}{0.85\linewidth}
    \tiny{A book was sold for 27.50 with a $10\%$ profit. If it's sold for 25.75, then what is the percent of profit and loss?}
    \vskip 5pt
    \begin{spacing}{0.5}
    \colorbox[RGB]{253,245,180}{\strut \tiny{To}}\colorbox[RGB]{253,245,180}{\strut \tiny{find}}\colorbox[RGB]{253,245,180}{\strut \tiny{the}}\colorbox[RGB]{253,245,180}{\strut \tiny{cost}}\colorbox[RGB]{253,245,180}{\strut \tiny{price}}\colorbox[RGB]{253,245,180}{\strut \tiny{(}}\colorbox[RGB]{253,245,180}{\strut \tiny{CP}}\colorbox[RGB]{253,245,180}{\strut \tiny{)}}\colorbox[RGB]{253,245,180}{\strut \tiny{of}}\colorbox[RGB]{253,245,180}{\strut \tiny{the}}\colorbox[RGB]{253,245,180}{\strut \tiny{book}}\colorbox[RGB]{253,244,179}{\strut \tiny{,}}\colorbox[RGB]{253,243,179}{\strut \tiny{we}}\colorbox[RGB]{253,245,180}{\strut \tiny{can}}\colorbox[RGB]{253,244,179}{\strut \tiny{use}}\colorbox[RGB]{253,245,180}{\strut \tiny{the}}\colorbox[RGB]{253,245,180}{\strut \tiny{formula}}\colorbox[RGB]{253,245,180}{\strut \tiny{:}}
    
    \colorbox[RGB]{253,245,180}{\strut \tiny{CP}}\colorbox[RGB]{253,244,179}{\strut \tiny{=}}\colorbox[RGB]{253,245,180}{\strut \tiny{Selling}}\colorbox[RGB]{253,245,180}{\strut \tiny{Price}}\colorbox[RGB]{253,245,180}{\strut \tiny{(}}\colorbox[RGB]{253,245,180}{\strut \tiny{SP}}\colorbox[RGB]{253,245,180}{\strut \tiny{)}}\colorbox[RGB]{253,245,180}{\strut \tiny{/}}\colorbox[RGB]{253,245,180}{\strut \tiny{(}}\colorbox[RGB]{253,245,180}{\strut \tiny{1}}\colorbox[RGB]{253,245,180}{\strut \tiny{+}}\colorbox[RGB]{253,245,180}{\strut \tiny{Profit}}\colorbox[RGB]{253,245,180}{\strut \tiny{Percentage}}\colorbox[RGB]{253,245,180}{\strut \tiny{)}}
    
    \colorbox[RGB]{253,245,180}{\strut \tiny{CP}}\colorbox[RGB]{253,245,180}{\strut \tiny{=}}\colorbox[RGB]{253,245,180}{\strut \tiny{27}}\colorbox[RGB]{253,245,180}{\strut \tiny{.}}\colorbox[RGB]{253,245,180}{\strut \tiny{50}}\colorbox[RGB]{253,245,180}{\strut \tiny{/}}\colorbox[RGB]{253,245,180}{\strut \tiny{(}}\colorbox[RGB]{253,245,180}{\strut \tiny{1}}\colorbox[RGB]{253,245,180}{\strut \tiny{+}}\colorbox[RGB]{253,245,180}{\strut \tiny{0}}\colorbox[RGB]{253,245,180}{\strut \tiny{.}}\colorbox[RGB]{253,245,180}{\strut \tiny{10}}\colorbox[RGB]{253,245,180}{\strut \tiny{)}}\colorbox[RGB]{253,245,180}{\strut \tiny{=}}\colorbox[RGB]{253,242,177}{\strut \tiny{25}}\colorbox[RGB]{253,245,180}{\strut \tiny{.}}\colorbox[RGB]{253,245,180}{\strut \tiny{00}}\colorbox[RGB]{253,245,180}{\strut \tiny{Now}}\colorbox[RGB]{253,245,180}{\strut \tiny{,}}\colorbox[RGB]{253,245,180}{\strut \tiny{we}}\colorbox[RGB]{253,245,180}{\strut \tiny{can}}\colorbox[RGB]{253,245,180}{\strut \tiny{calculate}}\colorbox[RGB]{253,245,180}{\strut \tiny{the}}\colorbox[RGB]{224,82,130}{\strut \tiny{loss}}\colorbox[RGB]{253,245,180}{\strut \tiny{as}}\colorbox[RGB]{253,245,180}{\strut \tiny{follows}}\colorbox[RGB]{253,245,180}{\strut \tiny{:}}
    
    \colorbox[RGB]{253,245,180}{\strut \tiny{Loss}}\colorbox[RGB]{253,245,180}{\strut \tiny{=}}\colorbox[RGB]{253,245,180}{\strut \tiny{CP}}\colorbox[RGB]{253,245,180}{\strut \tiny{-}}\colorbox[RGB]{253,245,180}{\strut \tiny{SP}}\colorbox[RGB]{253,245,180}{\strut \tiny{...}}\colorbox[RGB]{253,245,180}{\strut \tiny{The}}\colorbox[RGB]{253,245,180}{\strut \tiny{correct}}\colorbox[RGB]{253,245,180}{\strut \tiny{answer}}\colorbox[RGB]{253,245,180}{\strut \tiny{is}}\colorbox[RGB]{253,245,180}{\strut \tiny{D}}\colorbox[RGB]{253,245,180}{\strut \tiny{)}}\colorbox[RGB]{253,245,180}{\strut \tiny{3}}\colorbox[RGB]{253,245,180}{\strut \tiny{$\%$}}\colorbox[RGB]{253,245,180}{\strut \tiny{Loss}}\colorbox[RGB]{253,245,180}{\strut \tiny{.}}
    \end{spacing}
    \end{minipage}
    \caption{Comparison of the model outcome distribution $o_t$ (top) and steering success (bottom) across tokens. The outcome distribution and steering success have similar dynamics, with the same change points detected by the CPD algorithm (highlighted text).}
    \label{fig:main-result}
    \vspace{-15pt}
\end{figure}

The goal of this work is to study how the hidden representations in LLMs change over the course of in-context learning and next-word prediction, and to consider how model steering relates to token-level uncertainty. More specifically, we apply Forking Paths Analysis~\cite{bigelow2025forking} to estimate an LLM's certainty at each individual token in text generation. 
We then steer LLMs at different tokens and consider how uncertainty dynamics can predict which points an LLM can or cannot be successfully steered.
Finally, we investigate whether token-level uncertainty dynamics can be directly predicted from an LLM's hidden states.

\section{Estimating Outcome Distribution with Forking Paths Analysis}
\label{sec:fpa}

We build on the work of \citet{bigelow2025forking}, who present a method for estimating token-level uncertainty dynamics for black-box neural text generation. The goal of this approach is to understand how, for a given text completion sequence (a \textit{base path}), re-sampling at different tokens or token indices causes changes in the model's uncertainty.

Their method, Forking Paths Analysis, involves a few main steps: first, given an arbitrary prompt for long-form text generation, such as a multi-hop reasoning question and a chain-of-thought prompt, sample a single \textit{base path} completion $x^*$, along with token logit probabilities and top-N alternate token probabilities. In the second step, for each token index $t$ in the base path, concatenate all tokens in the base path up to $t$ (i.e. $x_{<t}^*)$ along with each top-N alternate token $w$ (i.e. $x_t = w$), and re-sample $S$ text completions from the LLM  $x^{(s)}_{>t}$ conditioned on this prompt. Using an answer extraction method, such as concatenating a string \textit{``Therefore, the answer is: \_\_\_''} and re-prompting the LLM, collect a final answer -- or \textit{outcome} -- as a one-hot vector $R(x)$ for each text completion. For the next step, collect these responses $R(x)$ and token probabilities $p(x_t = w  \ | \  x^*_{<t})$, and $p(x_{>t}^{(s)} \ | \ x^*_{<t}, x_t = w)$ into a weighted \textit{outcome distribution} $o_t$ for each token index $t$: $o_t = \mathbb{E}_{w, s} \big [  R(x^*_{<t}, \   x_t = w, \    x_{>t}^{(s)}) \big ]$

The \textit{outcome distribution} $o_t$ thus represents the LLM's uncertainty over final answers at each token index $t$. In the final step of Forking Paths Analysis, statistical models are used to analyze $o_t$ and look for specific token indexes $t$ where the outcome distribution changes suddenly. Bayesian Change Point Detection (CPD) models are used to estimate the probability $p(\tau = t \ | \ o_t)$ that a change point occurs at each token index $t$.
\citet{bigelow2025forking} analyze uncertainty dynamics of GPT-3.5 on a variety of common LLM benchmarks and find many striking cases of ``Forking Tokens'' where the outcome distribution suddenly shifts from one distribution of certainty to another. A major drawback of this approach, however, is the computational cost -- on the order of millions of tokens to analyze a single base path completion. In this paper, we investigate whether hidden states provide meaningful information about the outcome distribution during generation of reasoning text, without needing to generate new tokens.

\section{Steering Outcome Distribution with Hidden State Interventions}
\label{sec:steer}

Forking Paths Analysis reveals interesting uncertainty dynamics, where an LLM's uncertainty can dramatically change upon sampling a single token (see top of Figure \ref{fig:main-result}). In this work, we analyze how these dynamics are reflected in the hidden representations of an LLM. For instance, can we steer an LLM away from its path once it has decided on an answer? By intervening on linear subspaces in a model's hidden activations, we investigate whether steering success depends on the particular token where steering occurs, and how this relates to uncertainty dynamics in the model's output.

\textbf{Methods} We apply difference-in-means interventions as in \citet{marks2023geometry} to steer a model during generation. To steer our model towards a desired outcome $A$, we sample $n = 500$ generations whose final outcome is $A$ and $n$ generations whose final outcome is some other answer, denoted as $\overline A$. Let $\mathbf{h}_t^{(A)}$ be the list of hidden activations over all the generations leading to answer $A$ at token $t$, and $\mathbf{h}_t^{(\overline A)}$ be the list of activations over all generations leading to an alternate outcome. We set up a linear mean-mass probe \citep{marks2023geometry} to create a steering vector $s_t$.
\begin{equation*}
    s_t^{(A)} = \frac{1}{n} \sum \mathbf{h}_t^{(A)} - \frac{1}{n} \sum \mathbf{h}_t^{(\overline A)}
\end{equation*}

We select the token position $t$ that best separate between generations leading to $A$ and ones leading to $\overline A$. Specifically, we treat $s_t^{(A)}$ as a linear separator between the sets of activations and construct the classifier $S_t^{(A)}(x) = \mathtt{sigmoid}(\hat{s}_t^{(A)} \cdot x)$. We choose the token $t$ whose corresponding classifier $S_t^{(A)}$ has the highest classification accuracy on hidden activations from a held-out set of generations.

Difference-in-means steering vectors over later tokens and middle layers ($t = 200$, $l = 12$ for the example in Figure~\ref{fig:main-result}) achieve the highest classification accuracy ($\sim 80\%$). It is worth noting that these token positions do not correspond to change points in the model generation (as described in Section \ref{sec:fpa}) but instead to the end of the generation, when the final answer likely already appears in the generated text. 

Let $s^{(A)}$ be steering vector with the highest classification accuracy for answer $A$. At every token position $t$ in the base path, we apply difference-in-means steering by adding the pre-computed steering vector $s^{(A)}$ to the activation $h_t$ at token $t$ in the base path. We then sample $k = 10$ continuations from the intervened model, repeatedly adding $s^{(A)}$ to the activation at every generated token.
We measure steering success at each token $t$ as the number of times out of $k$ that the steered outcome is answer $A$, subtracted by the model's original outcome distribution $o_t(A)$ for $A$ at token $t$, i.e., the difference between steering success rate and base success rate without steering. High steering success corresponds to token positions at which we can control generation by intervening on linear subspaces in the model's activations.

\textbf{Data and Models} We analyze the \texttt{Llama-3.2 3B Instruct} LLM prompted with zero-shot chain-of-thought reasoning (i.e., prefixing ``Let's think step by step'' before the model's completion). Given the computational complexity of Forking Paths Analysis, we consider four examples from three reasoning datasets on which the LLM is uncertain. See Appendix \ref{app:data-and-examples} for more details.

\textbf{Results}  Our main steering analysis is presented in Figure \ref{fig:main-result}. In this example, sampling token $t = 114$ (``loss'') drastically shifts the outcome distribution away from ``B) 3\% profit'', which is the correct answer, and towards ``D) 3\% loss'', which is the model's generated answer. 

The success rate of steering in this case has very similar dynamics to the outcome distribution. Steering success is relatively high until $t = 114$, at which point it abruptly drops down to nearly zero. Given the illustrated example, we hypothesize that controllability (measured by steering success) correlates with uncertainty (measured by the outcome distribution $o_t$). That is, \textit{models are most steerable when they are least certain} about the final outcome.

\begin{figure}
    \centering
    \includegraphics[width=0.7\linewidth]{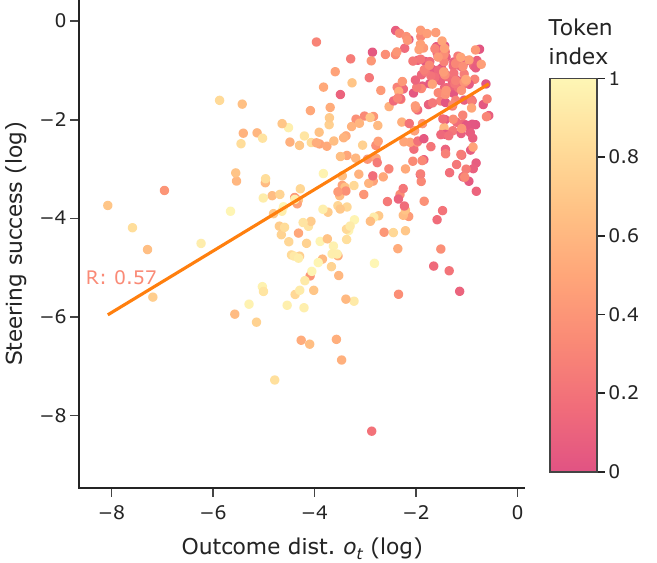}
    \caption{Correlation between steering success ($y$-axis) and base outcome probability ($x$-axis) across token positions.}
    \label{fig:control-vs-certainty}
    \vspace{-15pt}
\end{figure}

In Figure \ref{fig:control-vs-certainty}, we plot the steering success for the answer ``B) $3\%$ profit'' over the model's original outcome probability for answer B) across different time steps on a $\log$--$\log$ scale. We find a moderate correlation between steering success and base probability ($R = 0.57$), as shown in Figure \ref{fig:control-vs-certainty}. Broadly speaking, our results suggest that steering is most effective when the model is unsure about its final answer. Hence, steering success is a promising estimate of uncertainty during generation.

\textbf{Discussion} The success of difference-in-means steering depends on the model uncertainty during generation. We find a \textit{moderate correlation between controllability and uncertainty}, as measured by steering success and base outcome probabilities, respectively. Hence, uncertainty estimates may be indicative of times during generation at which to intervene on a model's activations to control its output. 

Estimating the uncertainty of a model during generation with Forking Paths Analysis is computationally expensive, since we must simulate the model's outcomes for branching continuations. Given that hidden state dynamics correlate with a model's uncertainty dynamics, can we efficiently predict the model's outcome distribution directly from its hidden states?

\section{Predicting Outcome Distribution from Hidden States}
\label{sec:probe}

\begin{figure}
    \centering
    \includegraphics[width=0.7\linewidth]{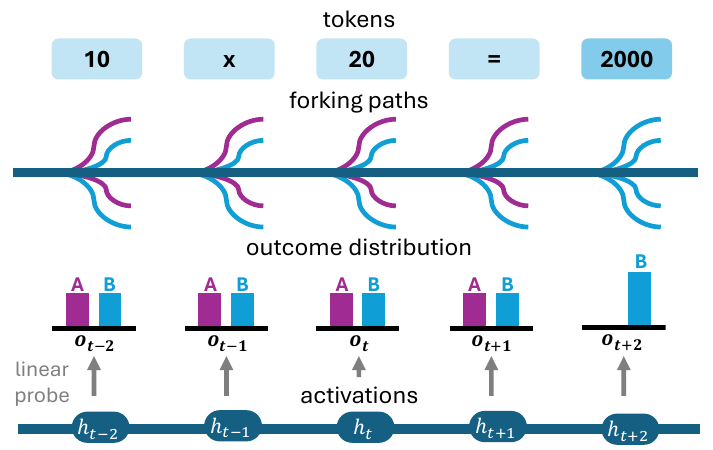}
    \caption{Our experimental set-up for Section \ref{sec:probe}. At every token position $t$, we train a linear probe to predict the distribution of outcomes $o_t$ from re-sampled paths starting at $t$, given the hidden representation $h_t$ over that token.}
    \label{fig:probe-main}
    \vspace{-10pt}
\end{figure}

Forking paths analysis estimates the distribution of outcomes $o_t$ by sampling alternate paths that diverge at token $t$. This is a computationally expensive process, because it must simulate the LLM's generation across numerous samples. However, research in faithfulness suggests that the reasoning chain leading to token $t$ is predictive of the final outcome distribution $o_t$ \cite{lanham2023measuring}. In this section, we try to estimate an LLM's outcome distribution $o_t$ from the activations over its reasoning process leading up to token $t$.

Is the semantic information in chain-of-thought text enough to predict a model's outcome distribution, or do the model's hidden activations contain information about its underlying decision-making that doesn't surface at the token level? 

We hypothesize that for a given token $t$, the embeddings $h_t$ from \texttt{Llama-3.2 3B Instruct} and $h'_t$ from \texttt{Gemma-2 2B Instruct} contain the same \textit{semantic information} about the chain-of-thought text up to token $t$, since they are both capable language models. However, only $h_t$ provides \textit{information about the underlying uncertainty dynamics} of the \texttt{Llama-3.2 3B Instruct} model.

We test our hypothesis by reporting KL losses for linear probes trained to predict the original model's outcome distribution from either the original model's or a separate model's hidden embeddings. Comparable KL losses indicate that model-specific hidden state dynamics aren't especially useful in predicting its outcome distribution (i.e., they are interchangeable with embeddings from a different LLM). Meanwhile, a lower KL loss for probes trained on the original model indicate that a model's hidden states are more predictive of its outcome distribution than the chain-of-thought text alone.

\textbf{Methods} Figure \ref{fig:probe-main} illustrates our experimental set-up. For each token index $t$ in the base path, consider the residual stream activation $h_t$ over that token. We train a linear probe to predict $o_t$ from $h_t$ over a set of token indices. Since $o_t$ is a full distribution, we train our linear probe with KL divergence loss and report the KL loss on a held-out validation set of token indices.

We also train a linear probe to predict the original model's outcome distribution $o_t$ from the text embedding $h'_t$ of a separate model, \texttt{Gemma-2 2B Instruct}, at token $t$. We report the probe's KL loss on the same validation set of token indices.

\textbf{Data and Models} We analyze \texttt{Llama-3.2 3B Instruct} as in Section \ref{sec:steer}. We report results on 10 randomly sampled examples from the AQuA dataset \cite{ling2017program}, including the example illustrated in Figure \ref{fig:main-result}.

\begin{figure}[b]
    \centering
    \includegraphics[width=0.6\linewidth]{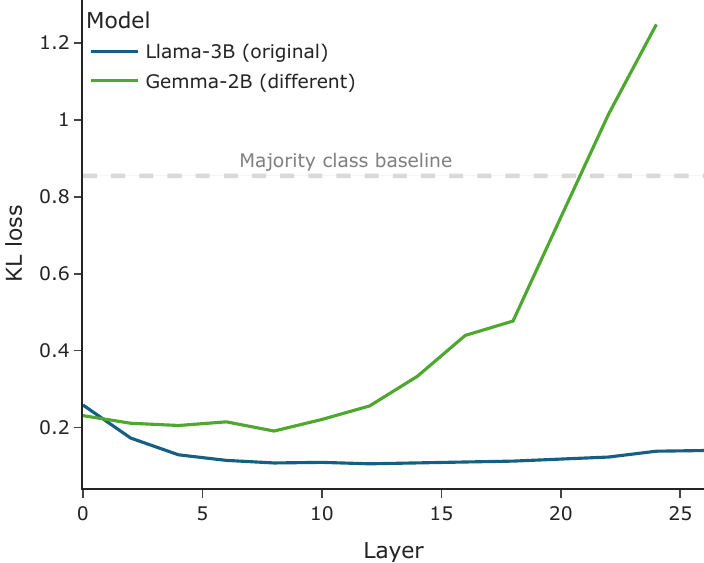}
    \caption{KL loss (\textit{lower is better}) for linear probes predicting the outcome distribution of \texttt{Llama} from the hidden representations of \texttt{Llama} (blue) and \texttt{Gemma} (green) at the same token mid-generation. Low loss suggests that hidden states over chain-of-thought text are predictive of \texttt{Llama}'s outcome distribution.}
    \label{fig:probe-future}
\end{figure}

\textbf{Results} Figure \ref{fig:probe-future} shows the KL loss for linear probes trained on $h_t$ and $h'_t$, averaged across our 10 data points. Both $h_t$ and $h'_t$ are more predictive than a random baseline which always predicts the uniform distribution ($1.53$ loss, not pictured) and a majority baseline which always predicts a one-hot distribution over the majority class ($0.85$ loss). Both probes are also most predictive around the middle layers of their respective model, around layers 6-10. This suggests that the reasoning chains have meaningful semantic information that's predictive of the model's outcome distribution.

The validation loss for the linear probe trained over the original activations is lower than the loss for the probe trained over activations from a different model ($0.11$ vs. $0.19$ at layer 8). The loss continues to be low for probes trained on the original model's later layers, while the loss rises for probes trained on the separate model's later layers. These results suggest that the original model's hidden activations $h_t$ may carry information that is used to determine the model's future actions, beyond the information carried by the output tokens alone.

\textbf{Discussion} The reasoning chain at token $t$ and its corresponding hidden activation, $h_t$, are both predictive of the model's outcome distribution $o_t$. However, hidden activations capture underlying decision-making information that is more predictive of the final outcome than the embedding of the same text by a different model. Probing a model's hidden activations is a promising direction for efficiently estimating its outcome distribution during generation.

\newpage
\bibliography{refs}

@article{fei2024nudging,
  title={Nudging: Inference-time Alignment via Model Collaboration},
  author={Fei, Yu and Razeghi, Yasaman and Singh, Sameer},
  journal={arXiv preprint arXiv:2410.09300},
  year={2024}
}

@article{bigelow2023context,
  title={In-Context Learning Dynamics with Random Binary Sequences},
  author={Bigelow, Eric J and Lubana, Ekdeep Singh and Dick, Robert P and Tanaka, Hidenori and Ullman, Tomer D},
  journal={International Conference on Learning Representations (ICLR)},
  year={2024}
}

@inproceedings{bigelow2025forking,
  title={Forking Paths in Neural Text Generation},
  author={Bigelow, Eric and Holtzman, Ari and Tanaka, Hidenori and Ullman, Tomer},
  booktitle={The Thirteenth International Conference on Learning Representations},
  year={2025}
}

@inproceedings{park2025iclr,
  title={ICLR: In-context learning of representations},
  author={Park, Core Francisco and Lee, Andrew and Lubana, Ekdeep Singh and Yang, Yongyi and Okawa, Maya and Nishi, Kento and Wattenberg, Martin and Tanaka, Hidenori},
  booktitle={The Thirteenth International Conference on Learning Representations},
  year={2025}
}

@article{marks2023geometry,
  title={The geometry of truth: Emergent linear structure in large language model representations of true/false datasets},
  author={Marks, Samuel and Tegmark, Max},
  journal={arXiv preprint arXiv:2310.06824},
  year={2023}
}

@inproceedings{ling2017program,
  title={Program Induction by Rationale Generation: Learning to Solve and Explain Algebraic Word Problems},
  author={Ling, Wang and Yogatama, Dani and Dyer, Chris and Blunsom, Phil},
  booktitle={Proceedings of the 55th Annual Meeting of the Association for Computational Linguistics (Volume 1: Long Papers)},
  pages={158--167},
  year={2017}
}

@article{kojima2022large,
  title={Large language models are zero-shot reasoners},
  author={Kojima, Takeshi and Gu, Shixiang Shane and Reid, Machel and Matsuo, Yutaka and Iwasawa, Yusuke},
  journal={Advances in neural information processing systems},
  volume={35},
  pages={22199--22213},
  year={2022}
}

@article{brown2020language,
  title={Language models are few-shot learners},
  author={Brown, Tom and Mann, Benjamin and Ryder, Nick and Subbiah, Melanie and Kaplan, Jared D and Dhariwal, Prafulla and Neelakantan, Arvind and Shyam, Pranav and Sastry, Girish and Askell, Amanda and others},
  journal={Advances in neural information processing systems},
  volume={33},
  pages={1877--1901},
  year={2020}
}

@article{xiong2024can,
  title={Can llms express their uncertainty? an empirical evaluation of confidence elicitation in llms},
  author={Xiong, Miao and Hu, Zhiyuan and Lu, Xinyang and Li, Yifei and Fu, Jie and He, Junxian and Hooi, Bryan},
  journal={International Conference on Learning Representations (ICLR)},
  year={2024}
}

@inproceedings{geng2024survey,
  title={A Survey of Confidence Estimation and Calibration in Large Language Models},
  author={Geng, Jiahui and Cai, Fengyu and Wang, Yuxia and Koeppl, Heinz and Nakov, Preslav and Gurevych, Iryna},
  booktitle={Proceedings of the 2024 Conference of the North American Chapter of the Association for Computational Linguistics: Human Language Technologies (Volume 1: Long Papers)},
  pages={6577--6595},
  year={2024}
}

@article{guo2025deepseek,
  title={Deepseek-r1: Incentivizing reasoning capability in llms via reinforcement learning},
  author={Guo, Daya and Yang, Dejian and Zhang, Haowei and Song, Junxiao and Zhang, Ruoyu and Xu, Runxin and Zhu, Qihao and Ma, Shirong and Wang, Peiyi and Bi, Xiao and others},
  journal={arXiv preprint arXiv:2501.12948},
  year={2025}
}

@article{jaech2024openai,
  title={Openai o1 system card},
  author={Jaech, Aaron and Kalai, Adam and Lerer, Adam and Richardson, Adam and El-Kishky, Ahmed and Low, Aiden and Helyar, Alec and Madry, Aleksander and Beutel, Alex and Carney, Alex and others},
  journal={arXiv preprint arXiv:2412.16720},
  year={2024}
}

@article{xie2021explanation,
  title={An explanation of in-context learning as implicit bayesian inference},
  author={Xie, Sang Michael and Raghunathan, Aditi and Liang, Percy and Ma, Tengyu},
  journal={arXiv preprint arXiv:2111.02080},
  year={2021}
}

@article{olsson2022context,
  title={In-context learning and induction heads},
  author={Olsson, Catherine and Elhage, Nelson and Nanda, Neel and Joseph, Nicholas and DasSarma, Nova and Henighan, Tom and Mann, Ben and Askell, Amanda and Bai, Yuntao and Chen, Anna and others},
  journal={arXiv preprint arXiv:2209.11895},
  year={2022}
}

@article{akyurek2022learning,
  title={What learning algorithm is in-context learning? investigations with linear models},
  author={Aky{\"u}rek, Ekin and Schuurmans, Dale and Andreas, Jacob and Ma, Tengyu and Zhou, Denny},
  journal={arXiv preprint arXiv:2211.15661},
  year={2022}
}

@article{arditi2024refusal,
  title={Refusal in language models is mediated by a single direction},
  author={Arditi, Andy and Obeso, Oscar and Syed, Aaquib and Paleka, Daniel and Panickssery, Nina and Gurnee, Wes and Nanda, Neel},
  journal={arXiv preprint arXiv:2406.11717},
  year={2024}
}

@article{ferrando2024primer,
  title={A primer on the inner workings of transformer-based language models},
  author={Ferrando, Javier and Sarti, Gabriele and Bisazza, Arianna and Costa-Juss{\`a}, Marta R},
  journal={arXiv preprint arXiv:2405.00208},
  year={2024}
}

@article{templeton2024scaling,
   title={Scaling Monosemanticity: Extracting Interpretable Features from Claude 3 Sonnet},
   author={Templeton, Adly and Conerly, Tom and Marcus, Jonathan and Lindsey, Jack and Bricken, Trenton and Chen, Brian and Pearce, Adam and Citro, Craig and Ameisen, Emmanuel and Jones, Andy and Cunningham, Hoagy and Turner, Nicholas L and McDougall, Callum and MacDiarmid, Monte and Freeman, C. Daniel and Sumers, Theodore R. and Rees, Edward and Batson, Joshua and Jermyn, Adam and Carter, Shan and Olah, Chris and Henighan, Tom},
   year={2024},
   journal={Transformer Circuits Thread},
   url={https://transformer-circuits.pub/2024/scaling-monosemanticity/index.html}
}

@article{panickssery2023steering,
  title={Steering llama 2 via contrastive activation addition},
  author={Panickssery, Nina and Gabrieli, Nick and Schulz, Julian and Tong, Meg and Hubinger, Evan and Turner, Alexander Matt},
  journal={arXiv preprint arXiv:2312.06681},
  year={2023}
}

@article{li2023inference,
  title={Inference-time intervention: Eliciting truthful answers from a language model},
  author={Li, Kenneth and Patel, Oam and Vi{\'e}gas, Fernanda and Pfister, Hanspeter and Wattenberg, Martin},
  journal={Advances in Neural Information Processing Systems},
  volume={36},
  pages={41451--41530},
  year={2023}
}

@article{turner2023steering,
  title={Steering language models with activation engineering},
  author={Turner, Alexander Matt and Thiergart, Lisa and Leech, Gavin and Udell, David and Vazquez, Juan J and Mini, Ulisse and MacDiarmid, Monte},
  journal={arXiv preprint arXiv:2308.10248},
  year={2023}
}

@inproceedings{rimsky2024steering,
  title={Steering Llama 2 via Contrastive Activation Addition},
  author={Rimsky, Nina and Gabrieli, Nick and Schulz, Julian and Tong, Meg and Hubinger, Evan and Turner, Alexander},
  booktitle={Proceedings of the 62nd Annual Meeting of the Association for Computational Linguistics (Volume 1: Long Papers)},
  pages={15504--15522},
  year={2024}
}

@article{cobbe2021gsm8k,
  title={Training Verifiers to Solve Math Word Problems},
  author={Cobbe, Karl and Kosaraju, Vineet and Bavarian, Mohammad and Chen, Mark and Jun, Heewoo and Kaiser, Lukasz and Plappert, Matthias and Tworek, Jerry and Hilton, Jacob and Nakano, Reiichiro and Hesse, Christopher and Schulman, John},
  journal={arXiv preprint arXiv:2110.14168},
  year={2021}
}

@inproceedings{rein2024gpqa,
  title={{GPQA}: A Graduate-Level Google-Proof Q\&A Benchmark},
  author={David Rein and Betty Li Hou and Asa Cooper Stickland and Jackson Petty and Richard Yuanzhe Pang and Julien Dirani and Julian Michael and Samuel R. Bowman},
  booktitle={First Conference on Language Modeling},
  year={2024},
  url={https://openreview.net/forum?id=Ti67584b98}
}

@article{lanham2023measuring,
  title={Measuring faithfulness in chain-of-thought reasoning},
  author={Lanham, Tamera and Chen, Anna and Radhakrishnan, Ansh and Steiner, Benoit and Denison, Carson and Hernandez, Danny and Li, Dustin and Durmus, Esin and Hubinger, Evan and Kernion, Jackson and others},
  journal={arXiv preprint arXiv:2307.13702},
  year={2023}
}
\bibliographystyle{icml2025}

\newpage
\appendix
\onecolumn

\section{Data Selection and Additional Examples}
\label{app:data-and-examples}

\begin{figure*}[ht]
    \centering
    \begin{subfigure}[t]{0.3\linewidth}
        \includegraphics[width=\linewidth]{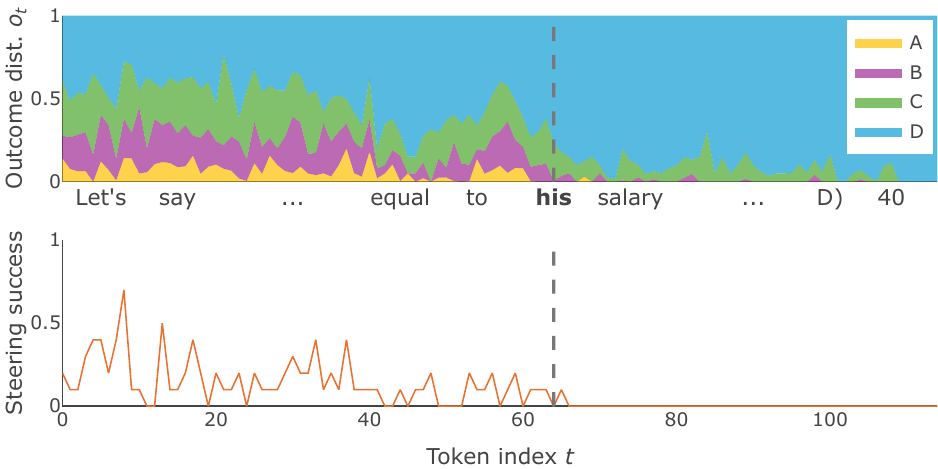}
        \tiny{The monthly salary S of a shop assistant is the sum of a fixed salary of \$500 plus 5\% of all monthly sales. What should the monthly sales be so that her monthly salary reaches \$1500?}
        \vspace{5pt}
        \begin{spacing}{0.5}
        \colorbox[RGB]{253,245,180}{\strut \tiny{Let}}\colorbox[RGB]{253,245,180}{\strut \tiny{'s}}\colorbox[RGB]{253,245,180}{\strut \tiny{say}}\colorbox[RGB]{253,245,180}{\strut \tiny{the}}\colorbox[RGB]{253,245,180}{\strut \tiny{sales}}\colorbox[RGB]{253,245,180}{\strut \tiny{person}}\colorbox[RGB]{253,245,180}{\strut \tiny{makes}}\colorbox[RGB]{253,245,180}{\strut \tiny{'}}\colorbox[RGB]{253,245,180}{\strut \tiny{x}}\colorbox[RGB]{253,241,176}{\strut \tiny{'}}\colorbox[RGB]{253,243,179}{\strut \tiny{sales}}\colorbox[RGB]{253,243,179}{\strut \tiny{.}}\colorbox[RGB]{253,244,179}{\strut \tiny{The}}\colorbox[RGB]{253,244,179}{\strut \tiny{total}}
        
        \colorbox[RGB]{253,241,176}{\strut \tiny{amount}}\colorbox[RGB]{253,244,179}{\strut \tiny{of}}\colorbox[RGB]{253,244,179}{\strut \tiny{money}}\colorbox[RGB]{253,244,179}{\strut \tiny{he}}\colorbox[RGB]{253,244,179}{\strut \tiny{makes}}\colorbox[RGB]{253,243,179}{\strut \tiny{from}}\colorbox[RGB]{253,242,177}{\strut \tiny{the}}\colorbox[RGB]{253,239,174}{\strut \tiny{sales}}\colorbox[RGB]{253,238,172}{\strut \tiny{is}}\colorbox[RGB]{253,229,163}{\strut \tiny{10}}\colorbox[RGB]{253,241,176}{\strut \tiny{$\%$}}\colorbox[RGB]{253,242,177}{\strut \tiny{of}}\colorbox[RGB]{253,242,177}{\strut \tiny{the}}
        
        \colorbox[RGB]{253,242,177}{\strut \tiny{total}}\colorbox[RGB]{253,242,177}{\strut \tiny{sales}}\colorbox[RGB]{253,242,177}{\strut \tiny{,}}\colorbox[RGB]{253,235,169}{\strut \tiny{which}}\colorbox[RGB]{253,243,179}{\strut \tiny{is}}\colorbox[RGB]{253,242,177}{\strut \tiny{0}}\colorbox[RGB]{253,238,173}{\strut \tiny{.}}\colorbox[RGB]{253,238,173}{\strut \tiny{1}}\colorbox[RGB]{253,242,177}{\strut \tiny{x}}\colorbox[RGB]{254,202,139}{\strut \tiny{*}}\colorbox[RGB]{253,240,175}{\strut \tiny{250}}\colorbox[RGB]{253,223,157}{\strut \tiny{=}}\colorbox[RGB]{253,240,175}{\strut \tiny{25}}\colorbox[RGB]{253,239,174}{\strut \tiny{x}}\colorbox[RGB]{253,239,174}{\strut \tiny{.}}\colorbox[RGB]{253,240,175}{\strut \tiny{The}}\colorbox[RGB]{253,243,179}{\strut \tiny{total}}
        
        \colorbox[RGB]{253,242,177}{\strut \tiny{amount}}\colorbox[RGB]{253,242,177}{\strut \tiny{of}}\colorbox[RGB]{253,243,178}{\strut \tiny{money}}\colorbox[RGB]{253,236,171}{\strut \tiny{he}}\colorbox[RGB]{253,233,167}{\strut \tiny{makes}}\colorbox[RGB]{253,233,168}{\strut \tiny{from}}\colorbox[RGB]{253,237,172}{\strut \tiny{the}}\colorbox[RGB]{253,240,175}{\strut \tiny{sales}}\colorbox[RGB]{253,237,172}{\strut \tiny{is}}\colorbox[RGB]{253,236,170}{\strut \tiny{also}}\colorbox[RGB]{253,243,179}{\strut \tiny{equal}}\colorbox[RGB]{253,243,178}{\strut \tiny{to}}
        
        \colorbox[RGB]{254,190,131}{\strut \tiny{his}}\colorbox[RGB]{253,223,157}{\strut \tiny{salary}}\colorbox[RGB]{253,240,175}{\strut \tiny{,}}\colorbox[RGB]{253,242,177}{\strut \tiny{which}}\colorbox[RGB]{253,242,177}{\strut \tiny{is}}\colorbox[RGB]{253,241,176}{\strut \tiny{\$}}\colorbox[RGB]{253,241,176}{\strut \tiny{100}}\colorbox[RGB]{253,240,175}{\strut \tiny{0}}\colorbox[RGB]{253,235,169}{\strut \tiny{.}}\colorbox[RGB]{253,242,177}{\strut \tiny{So}}\colorbox[RGB]{253,243,179}{\strut \tiny{,}}\colorbox[RGB]{253,243,179}{\strut \tiny{we}}\colorbox[RGB]{253,243,178}{\strut \tiny{can}}\colorbox[RGB]{253,243,178}{\strut \tiny{set}}\colorbox[RGB]{253,242,177}{\strut \tiny{up}}\colorbox[RGB]{253,242,177}{\strut \tiny{the}}
        
        \colorbox[RGB]{253,242,177}{\strut \tiny{equation}}\colorbox[RGB]{253,243,179}{\strut \tiny{:}}\colorbox[RGB]{253,240,175}{\strut \tiny{25}}\colorbox[RGB]{253,243,179}{\strut \tiny{x}}\colorbox[RGB]{253,244,179}{\strut \tiny{=}}\colorbox[RGB]{253,244,179}{\strut \tiny{100}}\colorbox[RGB]{253,244,179}{\strut \tiny{0}}\colorbox[RGB]{253,243,179}{\strut \tiny{.}}\colorbox[RGB]{253,244,179}{\strut \tiny{Div}}\colorbox[RGB]{253,245,180}{\strut \tiny{iding}}\colorbox[RGB]{253,245,180}{\strut \tiny{both}}\colorbox[RGB]{253,245,180}{\strut \tiny{sides}}\colorbox[RGB]{253,245,180}{\strut \tiny{by}}\colorbox[RGB]{253,245,180}{\strut \tiny{25}}\colorbox[RGB]{253,245,180}{\strut \tiny{,}}\colorbox[RGB]{253,245,180}{\strut \tiny{we}}
        
        \colorbox[RGB]{253,245,180}{\strut \tiny{get}}\colorbox[RGB]{253,245,180}{\strut \tiny{x}}\colorbox[RGB]{253,245,180}{\strut \tiny{=}}\colorbox[RGB]{253,245,180}{\strut \tiny{40}}\colorbox[RGB]{253,245,180}{\strut \tiny{.}}\colorbox[RGB]{253,245,180}{\strut \tiny{The}}\colorbox[RGB]{253,245,180}{\strut \tiny{answer}}\colorbox[RGB]{253,245,180}{\strut \tiny{is}}\colorbox[RGB]{253,245,180}{\strut \tiny{D}}\colorbox[RGB]{253,245,180}{\strut \tiny{)}}\colorbox[RGB]{253,245,180}{\strut \tiny{40}}\colorbox[RGB]{253,245,180}{\strut \tiny{.}}
        \end{spacing}
    \end{subfigure}
    \hfill
    \begin{subfigure}[t]{0.3\linewidth}
        \includegraphics[width=\linewidth]{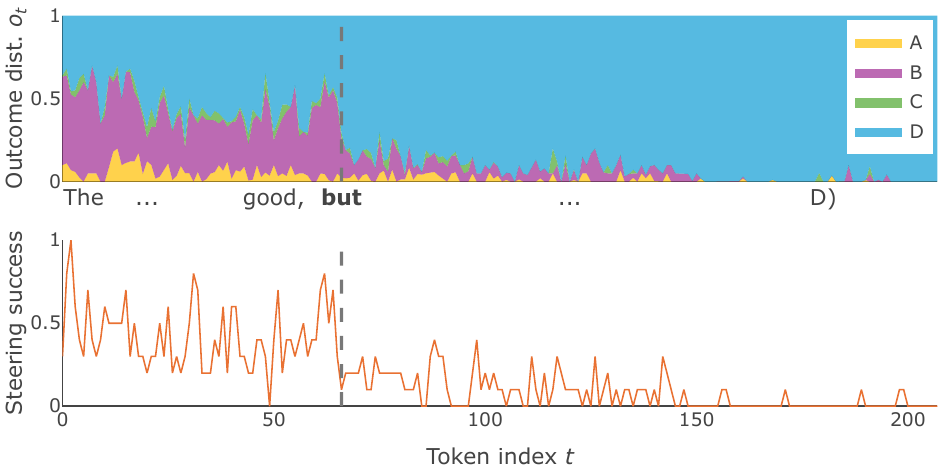}
        \tiny{There has been an outbreak of an viral infectious disease in your city. You have been asked to design a molecular diagnostic kit for quick detection of this retrovirus. How would you go about designing this kit?}
        \vspace{5pt}
        \begin{spacing}{0.5}
        \colorbox[RGB]{253,245,180}{\strut \tiny{The}}\colorbox[RGB]{253,245,180}{\strut \tiny{question}}\colorbox[RGB]{253,245,180}{\strut \tiny{is}}\colorbox[RGB]{253,245,180}{\strut \tiny{about}}\colorbox[RGB]{253,245,180}{\strut \tiny{designing}}\colorbox[RGB]{253,245,180}{\strut \tiny{a}}\colorbox[RGB]{253,245,180}{\strut \tiny{molecular}}\colorbox[RGB]{253,245,180}{\strut \tiny{diagnostic}}\colorbox[RGB]{253,245,180}{\strut \tiny{kit}}\colorbox[RGB]{253,230,164}{\strut \tiny{for}}
        
        \colorbox[RGB]{253,245,180}{\strut \tiny{a}}\colorbox[RGB]{253,245,180}{\strut \tiny{retro}}\colorbox[RGB]{253,245,180}{\strut \tiny{v}}\colorbox[RGB]{253,245,180}{\strut \tiny{irus}}\colorbox[RGB]{253,245,180}{\strut \tiny{.}}
        
        \colorbox[RGB]{253,226,160}{\strut \tiny{A}}\colorbox[RGB]{253,229,163}{\strut \tiny{)}}\colorbox[RGB]{253,230,164}{\strut \tiny{This}}\colorbox[RGB]{253,244,179}{\strut \tiny{option}}\colorbox[RGB]{253,243,178}{\strut \tiny{is}}\colorbox[RGB]{253,240,175}{\strut \tiny{not}}\colorbox[RGB]{253,243,179}{\strut \tiny{ideal}}\colorbox[RGB]{253,242,177}{\strut \tiny{because}}\colorbox[RGB]{253,240,175}{\strut \tiny{...}}
        
        \colorbox[RGB]{253,245,180}{\strut \tiny{B}}\colorbox[RGB]{253,245,180}{\strut \tiny{)}}\colorbox[RGB]{253,245,180}{\strut \tiny{This}}\colorbox[RGB]{253,245,180}{\strut \tiny{option}}\colorbox[RGB]{253,245,180}{\strut \tiny{is}}\colorbox[RGB]{253,245,180}{\strut \tiny{a}}\colorbox[RGB]{253,245,180}{\strut \tiny{good}}\colorbox[RGB]{253,245,180}{\strut \tiny{approach}}\colorbox[RGB]{253,245,180}{\strut \tiny{,}}\colorbox[RGB]{235,101,123}{\strut \tiny{but}}\colorbox[RGB]{253,227,161}{\strut \tiny{...}}
        
        \colorbox[RGB]{253,244,179}{\strut \tiny{C}}\colorbox[RGB]{253,244,179}{\strut \tiny{)}}\colorbox[RGB]{253,244,179}{\strut \tiny{This}}\colorbox[RGB]{253,244,179}{\strut \tiny{option}}\colorbox[RGB]{253,244,179}{\strut \tiny{is}}\colorbox[RGB]{253,244,179}{\strut \tiny{not}}\colorbox[RGB]{253,244,179}{\strut \tiny{ideal}}\colorbox[RGB]{253,244,179}{\strut \tiny{because}}\colorbox[RGB]{253,244,179}{\strut \tiny{...}}
        
        \colorbox[RGB]{253,243,179}{\strut \tiny{D}}\colorbox[RGB]{253,243,179}{\strut \tiny{)}}\colorbox[RGB]{253,243,179}{\strut \tiny{This}}\colorbox[RGB]{253,243,179}{\strut \tiny{option}}\colorbox[RGB]{253,243,179}{\strut \tiny{is}}\colorbox[RGB]{253,243,179}{\strut \tiny{the}}\colorbox[RGB]{253,243,179}{\strut \tiny{best}}\colorbox[RGB]{253,243,179}{\strut \tiny{approach}}\colorbox[RGB]{253,243,179}{\strut \tiny{...}}

        \colorbox[RGB]{253,245,180}{\strut \tiny{Therefore}}\colorbox[RGB]{253,245,180}{\strut \tiny{,}}\colorbox[RGB]{253,245,180}{\strut \tiny{the}}\colorbox[RGB]{253,245,180}{\strut \tiny{best}}\colorbox[RGB]{253,245,180}{\strut \tiny{answer}}\colorbox[RGB]{253,245,180}{\strut \tiny{is}}\colorbox[RGB]{253,245,180}{\strut \tiny{D}}\colorbox[RGB]{253,245,180}{\strut \tiny{.}}
        \end{spacing}
    \end{subfigure}
    \hfill
    \begin{subfigure}[t]{0.3\linewidth}
        \includegraphics[width=\linewidth]{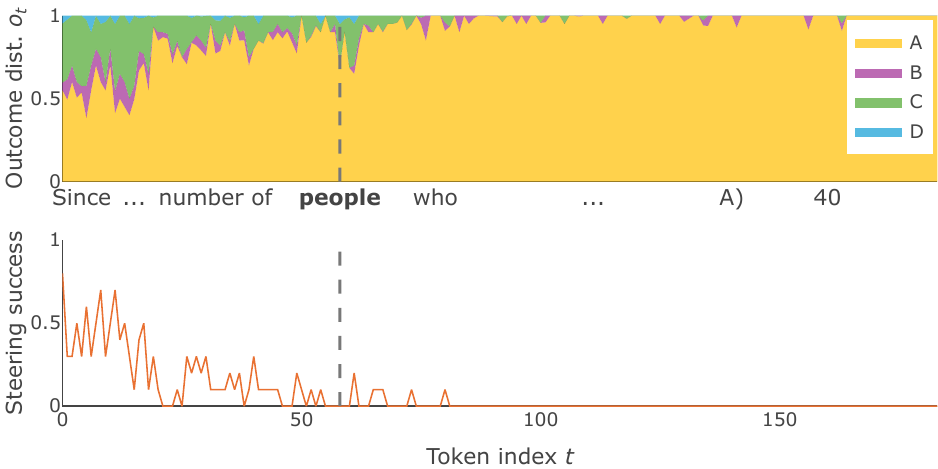}
        \tiny{Courtney attended a concert and reported that the audience was 48 in number. However, Kelly went to the same concert and said that Courtney had made the mistake of overstating the number of people in attendance by 20\%. If Kelly was right, how many people really attended the concert?}
        \vspace{5pt}
        \begin{spacing}{0.5}
        \colorbox[RGB]{253,245,180}{\strut \tiny{Since}}\colorbox[RGB]{253,245,180}{\strut \tiny{Kelly}}\colorbox[RGB]{253,245,180}{\strut \tiny{said}}\colorbox[RGB]{253,245,180}{\strut \tiny{Courtney}}\colorbox[RGB]{253,245,180}{\strut \tiny{overst}}\colorbox[RGB]{253,245,180}{\strut \tiny{ated}}\colorbox[RGB]{253,245,180}{\strut \tiny{the}}\colorbox[RGB]{253,245,180}{\strut \tiny{number}}\colorbox[RGB]{253,245,180}{\strut \tiny{of}}\colorbox[RGB]{253,245,180}{\strut \tiny{people}}
        
        \colorbox[RGB]{253,245,180}{\strut \tiny{by}}\colorbox[RGB]{253,245,180}{\strut \tiny{20}}\colorbox[RGB]{253,245,180}{\strut \tiny{$\%$,}}\colorbox[RGB]{253,245,180}{\strut \tiny{we}}\colorbox[RGB]{253,245,180}{\strut \tiny{can}}\colorbox[RGB]{253,244,179}{\strut \tiny{set}}\colorbox[RGB]{230,92,127}{\strut \tiny{up}}\colorbox[RGB]{253,245,180}{\strut \tiny{an}}\colorbox[RGB]{253,245,180}{\strut \tiny{equation}}\colorbox[RGB]{253,242,177}{\strut \tiny{.}}\colorbox[RGB]{253,244,179}{\strut \tiny{Courtney}}\colorbox[RGB]{253,245,180}{\strut \tiny{said}}
        
        \colorbox[RGB]{253,245,180}{\strut \tiny{there}}\colorbox[RGB]{253,245,180}{\strut \tiny{were}}\colorbox[RGB]{253,242,177}{\strut \tiny{48}}\colorbox[RGB]{232,96,125}{\strut \tiny{people}}\colorbox[RGB]{253,245,180}{\strut \tiny{,}}\colorbox[RGB]{253,242,177}{\strut \tiny{but}}\colorbox[RGB]{253,245,180}{\strut \tiny{Kelly}}\colorbox[RGB]{253,243,179}{\strut \tiny{said}}\colorbox[RGB]{253,239,174}{\strut \tiny{this}}\colorbox[RGB]{253,245,180}{\strut \tiny{was}}\colorbox[RGB]{253,245,180}{\strut \tiny{20}}\colorbox[RGB]{253,245,180}{\strut \tiny{$\%$}}\colorbox[RGB]{253,245,180}{\strut \tiny{more}}
        
        \colorbox[RGB]{253,221,154}{\strut \tiny{than}}\colorbox[RGB]{254,219,153}{\strut \tiny{the}}\colorbox[RGB]{253,220,153}{\strut \tiny{actual}}\colorbox[RGB]{254,214,149}{\strut \tiny{number}}\colorbox[RGB]{253,234,168}{\strut \tiny{.}}\colorbox[RGB]{253,238,172}{\strut \tiny{So}}\colorbox[RGB]{253,238,172}{\strut \tiny{,}}\colorbox[RGB]{253,240,175}{\strut \tiny{we}}\colorbox[RGB]{253,243,178}{\strut \tiny{can}}\colorbox[RGB]{253,242,177}{\strut \tiny{set}}\colorbox[RGB]{253,243,178}{\strut \tiny{up}}\colorbox[RGB]{253,242,177}{\strut \tiny{the}}\colorbox[RGB]{253,245,180}{\strut \tiny{equation}}\colorbox[RGB]{253,245,180}{\strut \tiny{:}}
        
        \colorbox[RGB]{253,245,180}{\strut \tiny{48}}\colorbox[RGB]{253,245,180}{\strut \tiny{=}}\colorbox[RGB]{253,245,180}{\strut \tiny{x}}\colorbox[RGB]{253,245,180}{\strut \tiny{+}}\colorbox[RGB]{253,245,180}{\strut \tiny{20}}\colorbox[RGB]{253,245,180}{\strut \tiny{$\%$}}\colorbox[RGB]{253,245,180}{\strut \tiny{of}}\colorbox[RGB]{253,245,180}{\strut \tiny{x}}\colorbox[RGB]{253,245,180}{\strut \tiny{To}}\colorbox[RGB]{253,245,180}{\strut \tiny{find}}\colorbox[RGB]{253,245,180}{\strut \tiny{20}}\colorbox[RGB]{253,245,180}{\strut \tiny{$\%$}}\colorbox[RGB]{253,245,180}{\strut \tiny{of}}\colorbox[RGB]{253,245,180}{\strut \tiny{x}}\colorbox[RGB]{253,245,180}{\strut \tiny{,}}\colorbox[RGB]{253,245,180}{\strut \tiny{we}}\colorbox[RGB]{253,245,180}{\strut \tiny{multiply}}\colorbox[RGB]{253,245,180}{\strut \tiny{2}}\colorbox[RGB]{253,245,180}{\strut \tiny{:}}
        
        \colorbox[RGB]{253,245,180}{\strut \tiny{48}}\colorbox[RGB]{253,245,180}{\strut \tiny{=}}\colorbox[RGB]{253,245,180}{\strut \tiny{x}}\colorbox[RGB]{253,245,180}{\strut \tiny{+}}\colorbox[RGB]{253,245,180}{\strut \tiny{0}}\colorbox[RGB]{253,245,180}{\strut \tiny{.}}\colorbox[RGB]{253,245,180}{\strut \tiny{2}}\colorbox[RGB]{253,245,180}{\strut \tiny{x}}\colorbox[RGB]{253,245,180}{\strut \tiny{48}}\colorbox[RGB]{253,245,180}{\strut \tiny{=}}\colorbox[RGB]{253,245,180}{\strut \tiny{1}}\colorbox[RGB]{253,245,180}{\strut \tiny{.}}\colorbox[RGB]{253,245,180}{\strut \tiny{2}}\colorbox[RGB]{253,245,180}{\strut \tiny{x}}\colorbox[RGB]{253,245,180}{\strut \tiny{Now}}\colorbox[RGB]{253,245,180}{\strut \tiny{,}}\colorbox[RGB]{253,245,180}{\strut \tiny{we}}\colorbox[RGB]{253,245,180}{\strut \tiny{can}}\colorbox[RGB]{253,245,180}{\strut \tiny{divide}}
        
        \colorbox[RGB]{253,245,180}{\strut \tiny{to}}\colorbox[RGB]{253,245,180}{\strut \tiny{solve}}\colorbox[RGB]{253,245,180}{\strut \tiny{for}}\colorbox[RGB]{253,245,180}{\strut \tiny{x}}\colorbox[RGB]{253,245,180}{\strut \tiny{:}}\colorbox[RGB]{253,245,180}{\strut \tiny{x}}\colorbox[RGB]{253,245,180}{\strut \tiny{=}}\colorbox[RGB]{253,245,180}{\strut \tiny{48}}\colorbox[RGB]{253,245,180}{\strut \tiny{/}}\colorbox[RGB]{253,245,180}{\strut \tiny{1}}\colorbox[RGB]{253,245,180}{\strut \tiny{.}}\colorbox[RGB]{253,245,180}{\strut \tiny{2}}\colorbox[RGB]{253,245,180}{\strut \tiny{x}}\colorbox[RGB]{253,245,180}{\strut \tiny{=}}\colorbox[RGB]{253,245,180}{\strut \tiny{40}}
        
        \colorbox[RGB]{253,245,180}{\strut \tiny{The}}\colorbox[RGB]{253,245,180}{\strut \tiny{correct}}\colorbox[RGB]{253,245,180}{\strut \tiny{answer}}\colorbox[RGB]{253,245,180}{\strut \tiny{is}}\colorbox[RGB]{253,245,180}{\strut \tiny{A}}\colorbox[RGB]{253,245,180}{\strut \tiny{)}}\colorbox[RGB]{253,245,180}{\strut \tiny{40}}\colorbox[RGB]{253,245,180}{\strut \tiny{.}}
        \end{spacing}
    \end{subfigure}
    \caption{Three additional examples of steering analysis. Each column corresponds to a single example, with the following structure: (Top) outcome distribution $o_t$ across token positions, estimated by re-sampling completions at alternate tokens for each token index (see Section \ref{sec:fpa}). (Middle) Steering success across token positions, estimated by the number of times a steering vector successfully changes the model's final answer (see Section \ref{sec:steer}). (Bottom) answer generated with greedy sampling, with highlighted change points in the outcome distribution $o_t$ (see Section \ref{sec:fpa}). Across all examples, the outcome distribution and steering success display similar dynamics, with a sharp shift at the highlighted change points.}
    \label{fig:extra-examples}
\end{figure*}

Due to the computational complexity of Forking Paths Analysis, we consider a handful of examples from the multiple-choice reasoning datasets: \textbf{GSM8k}, a collection of grade-school math questions \citep{cobbe2021gsm8k}; \textbf{AQuA}, a collection of algebraic word problems \citep{ling2017program}; and \textbf{GPQA}, a collection of graduate-level science questions \citep{rein2024gpqa}.

We select our data points by sampling from each dataset and keeping questions for which the LLM we analyze (\texttt{Llama-3.2 3B Instruct}) is uncertain. In particular, we sample $k = 10$ generations from the LLM, and extract the answer from each generation by prompting the model at the end with ``What is your final answer?''. We choose examples where the frequency of the most common answer across the 10 examples is between 4 and 6. That is, the LLM is only $40$--$60\%$ likely to generate that answer. Figure \ref{fig:extra-examples} shows the uncertainty and steering dynamics for our selected examples. The steering correlations reported in Section \ref{sec:steer} pertain to the example in Figure \ref{fig:main-result} in the main text. The correlation coefficient for steering success and outcome distribution (on a $\log$--$\log$ scale as in Figure \ref{fig:control-vs-certainty}) averaged across our four examples is $R = 0.64$.

\end{document}